%% file: main.tex
\documentclass[runningheads]{llncs}
\usepackage[T1]{fontenc}
\usepackage{graphicx}
\usepackage{booktabs}
\usepackage[misc]{ifsym}
\newcommand{\corr}{(\Letter)}
\usepackage{times}
\usepackage{soul}
\usepackage{url}
\usepackage[hidelinks]{hyperref}
\usepackage[utf8]{inputenc}
\usepackage[small]{caption}
\usepackage{amsfonts,amssymb,amsmath,bm}
\usepackage[switch]{lineno}
\usepackage{textcomp}
\usepackage{xcolor}
\usepackage{subfigure}
\usepackage{multirow}
\usepackage{booktabs}
\usepackage[ruled,linesnumbered]{algorithm2e}
\usepackage{url}
\usepackage{setspace}
\usepackage{caption}
\usepackage{CJK}
\usepackage{enumitem}

\usepackage{placeins}

\newcommand{\ie}{\textit{i.e.}}

\begin{document}

\title{AdvKT: An Adversarial Multi-Step Training Framework for Knowledge Tracing}

\titlerunning{AdvKT: An Adversarial Multi-Step Training Framework for Knowledge Tracing}

\toctitle{AdvKT: An Adversarial Multi-Step Training Framework for Knowledge Tracing}  
\tocauthor{Lingyue Fu, Ting Long, Jianghao Lin, Wei Xia, Xinyi Dai, Ruiming Tang, Yasheng Wang, Weinan Zhang, Yong Yu} 

\author{Lingyue Fu\inst{1} \and Ting Long\inst{2} \and Jianghao Lin\inst{1} \and Wei Xia\inst{4} \and Xinyi Dai\inst{3} \and Ruiming Tang\inst{3} \and  
Yasheng Wang\inst{3} \and
Weinan Zhang\inst{1}\corr \and
Yong Yu\inst{1}\corr}

\authorrunning{L. Fu et al.}
\institute{Shanghai Jiao Tong University \\ \email{\{fulingyue, chiangel, wnzhang, yyu\}@sjtu.edu.cn}
\and 
Jilin University \email{longting@jlu.edu.cn}
\and
 Huawei Noah's Ark Lab 
 \\ \email{\{daixinyi5, tangruiming, wangyasheng\}@huawei.com}
 \and
 www.imxwell.com
 \email{xwell.xia@gmail.com}
 }


\maketitle              

\begin{abstract}
Knowledge Tracing (KT) monitors students' knowledge states and simulates their responses to question sequences. Existing KT models typically follow a single-step training paradigm, which leads to discrepancies with the multi-step inference process required in real-world simulations, resulting in significant error accumulation. This accumulation of error, coupled with the issue of data sparsity, can substantially degrade the performance of recommendation models in the intelligent tutoring systems. To address these challenges, we propose a novel Adversarial Multi-Step Training Framework for Knowledge Tracing (AdvKT), which, for the first time, focuses on the multi-step KT task. More specifically, AdvKT leverages adversarial learning paradigm involving a generator and a discriminator. The generator mimics high-reward responses, effectively reducing error accumulation across multiple steps, while the discriminator provides feedback to generate synthetic data. Additionally, we design specialized data augmentation techniques to enrich the training data with realistic variations, ensuring that the model generalizes well even in scenarios with sparse data.
Experiments conducted on four real-world datasets demonstrate the superiority of AdvKT over existing KT models, showcasing its ability to address both error accumulation and data sparsity issues effectively.

\end{abstract}

\section{Introduction}
Intelligent tutoring systems (ITS), such as Massive Online Open Courses (MOOCs), aim to help students learn more efficiently through learning path recommendation and question recommendation. 
Knowledge Tracing (KT) is a vital component of ITS, which 
estimates the knowledge state by predicting the student responses, i.e., whether a student can answer each question correctly or not. 
KT models are used in two ways: as monitors to provide human instructors with insights into students’ mastery levels, and as \textit{student simulators} to supply reward signals for recommendation models.

\begin{figure}[tb]
\centering
    \subfigure[Application of the student simulator.]{
    \includegraphics[width=0.7\linewidth]{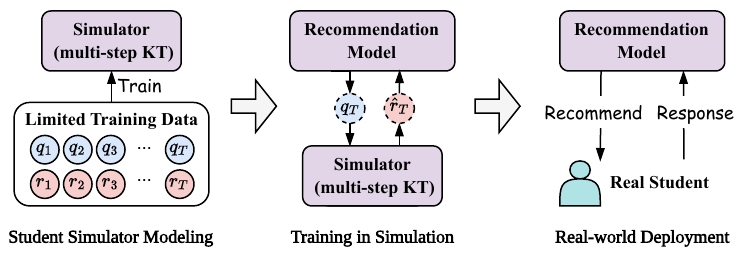}
    \label{demo-its}
    }
    \\
    \subfigure[Inference Comparison.]{
        \includegraphics[width=0.27\linewidth]{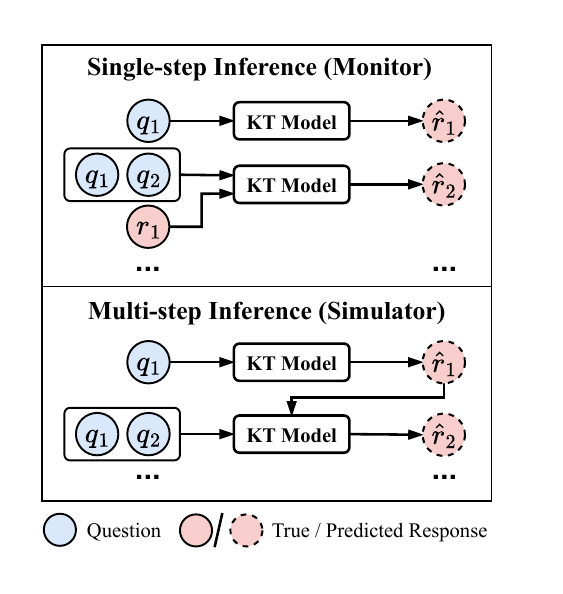}
         \label{fig:demo}
    }
    \subfigure[AUC Comparison.]{
        \includegraphics[width=0.31\linewidth]{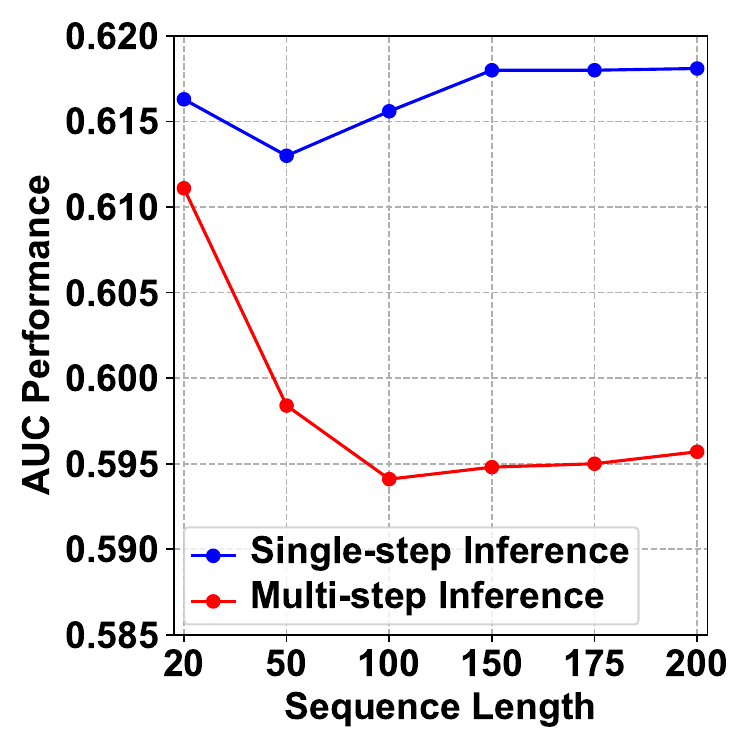}
    \label{fig:datacom}
    }
    \caption{(a) Illustration of the students simulator's application and the comparison between single-step and multi-step inference KT. (b) Comparison of single-step inference and multi-step inference (simulator) and the primary application of the simulator. Input of single-step inference: historical question and ground-truth response. Input of multi-step inference: historical question and predicted response. (c) Performance comparison of the single-step inference and multi-step inference with a single-step training approach.
    With increased sequence length, errors accumulate gradually under the multi-step inference setting.}
    \label{fig:singlevsmulti}
\end{figure}

Existing KT models~\cite{SKVMN,DKT,AKT,simplekt,stableKT} show promising performance as monitors of students’ learning progress. By taking students' real-time responses to various questions as input, these models estimate the knowledge states of students and make predictions for the next questions. The predictions can be a reference for human instructors, helping them identify areas where students may need additional support or challenge.

KT models are more frequently employed in ITS as \textit{student simulators} for question recommendation models~\cite{ECSAL,gong2020attentional,li2023graph}. As illustrated in Figure \ref{demo-its}, after the recommendation model generates a sequence of questions, the pre-trained simulator (KT model) provides simulated student responses, which serve as rewards to optimize the recommendation model. The objective of the recommendation model is to maximize the reward provided by the student simulator. In this setting, the accuracy of the simulation has a significant impact on the effectiveness of the recommendation model in the real-world deployment phase. 

Despite their widespread use, existing KT models largely overlook their role as student simulators, leading to a gap between \textit{single-step training} and \textit{multi-step inference}. Typically, these models follow a single-step training paradigm, where they are optimized to predict a student’s response for only the next question in a sequence, assuming the ground-truth response history is always accessible. In real-world simulations, however, the simulator must predict responses for an entire recommended question sequence at once. As illustrated in Figure~\ref{fig:demo}, during multi-step inference, the simulator iteratively generates responses based on its previous predictions, as ground-truth history is unavailable.

The discrepancy between single-step training and multi-step inference results in significant \textit{error accumulation} over longer sequences, adversely affecting the recommendation model's performance. As shown in Figure~\ref{fig:datacom}, we train EERNNA~\cite{EERNNA} using a single-step training approach and evaluate it under both single-step and multi-step inference settings. In the single-step setting, where ground-truth history is available, the model performs stably across varying sequence lengths, as input sequences remain error-free. In contrast, under the multi-step inference setting, prediction errors gradually accumulate as the sequence length increases, leading to a substantial degradation in performance. This discrepancy highlights the need for KT models that better align with the multi-step inference nature of real-world simulations.

Furthermore, current ITS suffer from the data sparsity problem.
Existing KT models are generally supervised by optimizing the binary cross-entropy (BCE) loss and, therefore, minimizing the KL divergence between the model and data distributions, which casts heavy demands on the quantity of training data.
However, in real-world ITS, the number of students in each class is typically limited, and each student exercises a finite number of questions. 
For example, 
in ASSIST09 dataset\footnote{https://sites.google.com/site/assistmentsdata/home/2009-2010-assistment-data}, there are only 2,661 students with 165,455 interactions, which contains 14,083 questions.
The interactive data are sparse with respect to the whole space of question sequences, thus it is difficult for KT models to capture the students' learning patterns precisely.


To conquer the above challenges, we propose a novel \underline{Adv}ersarial Multi-Step Training Framework for \underline{K}nowledge \underline{T}racing, named \textbf{AdvKT}.
To the best of our knowledge, this is the first work to formalize the multi-step KT task and the first to introduce adversarial learning to the KT task.
AdvKT enables models to effectively handle the iterative nature of multi-step simulations by leveraging adversarial learning to diversify training sequences, thereby mitigating error propagation during inference. 
Additionally, we propose specialized data augmentation techniques to enrich training data with realistic variations, ensuring the model generalizes effectively even in sparse data scenarios.

To sum up, our contributions are summarized as follows:
\begin{itemize} [leftmargin=10pt]
    \item We propose a novel Adversarial Multi-Step Training Framework for the KT task (\ie, \textbf{AdvKT}) to address the challenge of error accumulation in multi-step inference.
    To the best of our knowledge, this is the first work that focuses on the nature of multi-step prediction in the KT task. 
    \item We design tailored data augmentation techniques specifically for KT tasks and incorporate them into AdvKT, thereby reducing the prediction bias caused by the data sparsity problem and uneven distributions.
    \item  Extensive experiments on four real-world datasets demonstrate that AdvKT achieves state-of-the-art performance under the multi-step inference setting, as well as superior capabilities to tackle the data sparsity.
\end{itemize}

\section{Problem Formulation}\label{sec:prob-statement}

In ITS, let $\mathcal{S}$ represent the set of students, $\mathcal{Q}$ the set of questions, and $\mathcal{C}$ the set of concepts. Each question $q \in \mathcal{Q}$ is associated with a subset of concepts from the set $\mathcal{C}$, denoted as $\mathcal{C}^q = \{c^q_1, \dots, c^q_k\} \subset \mathcal{C}$. The student's learning sequence is recorded as $s = [(q_i, r_i)]_{i=1}^{T}$, where $q_i \in \mathcal{Q}$ is the question, and $r_i \in {0,1}$ indicates whether the student answered question $q_i$ correctly ($r_i = 1$ for correct, and $r_i = 0$ for incorrect).

Previous KT models typically follow a single-step training and inference paradigm, where the ground-truth responses to previous questions are used to predict the next response. The single-step prediction at step $T$ can be formulated as:
\begin{equation}
    \hat{r}_T = p\left(r_T=1|[(q_t,r_t)]_{t=1}^{T-1}, q_T\right).
\end{equation}

In contrast to existing KT models, we focus on a multi-step prediction paradigm, which is more in line with real-world applications. Specifically, this approach involves predicting the student's response $r_T$ using previously predicted responses $[\hat{r}_t]_{t=1}^{T-1}$, which can be expressed as:
\begin{equation}
    \hat r_{T} = p\left(r_T = 1|[(q_t, \hat r_t)]_{t=1}^{T-1}, q_{T}\right).
\end{equation}

\section{Methodology}
\begin{figure*}[t]
    \centering
    \includegraphics[width=\textwidth]{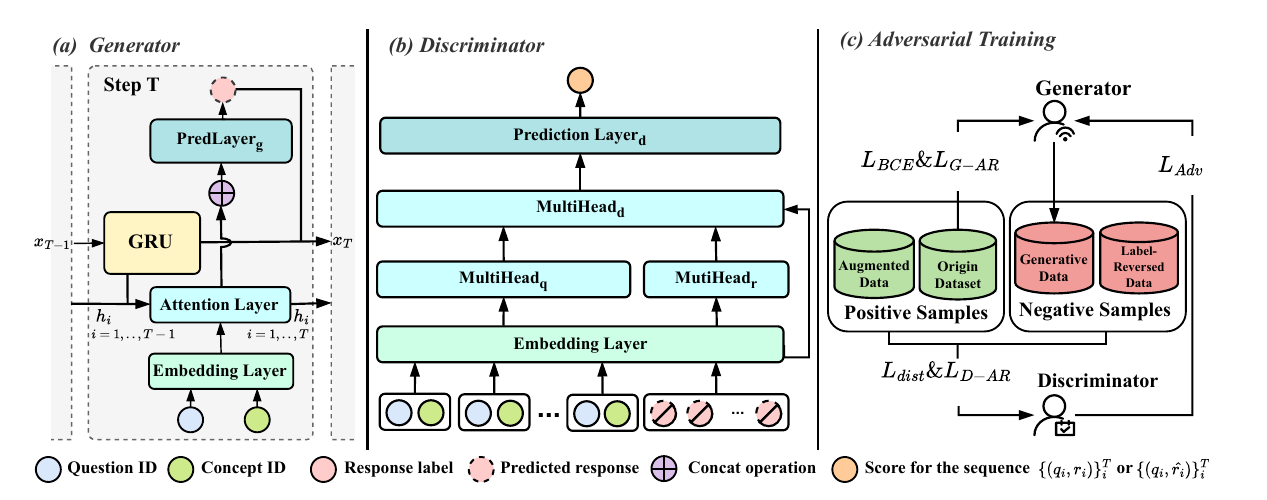}
    \caption{The overview of the proposed AdvKT framework. }
    \label{fig:framework}
\end{figure*}

In this section, we introduce the framework of AdvKT, as dipicted in Figure~\ref{fig:framework}.
AdvKT consists of two main components: a generator that predicts the student's responses for a given question sequence, and a discriminator that distinguishes between real and fake data.
We employ adversarial training over the generator and the discriminator with data augmentation.

\subsection{Embedding Layer}
To project discrete IDs to dense representations, we create a question embedding matrix $\textbf{Emb}^Q\in \mathbb{R}^{|\mathcal{Q}|\times d}$ and a concept embedding matrix $\textbf{Emb}^C\in \mathbb{R}^{|\mathcal{C}|\times d}$, where $d$ denotes the embedding size.
Let $\mathbf{e}_q = \textbf{Emb}^Q_{q} \in \mathbb{R}^d$ and $\mathbf{e}_c = \textbf{Emb}^C_c \in \mathbb{R}^d$ be the embedding vectors of question $q$ and concept $c$, respectively.
Following~\cite{DHKT}, we represent question $q$ by concatenating the question embedding with the mean aggregation of its corresponding concept embeddings:
\begin{equation}
\begin{aligned}
    \mathbf{v}_{c,q}&=\frac{1}{|\mathcal{C}^q|} \sum\nolimits_{i=1}^{|\mathcal{C}^q|} \mathbf{e}_{c_i^q}, 
    \mathbf{v}_{q} &= \mathbf{e}_q \oplus \mathbf{v}_{c,q},
\end{aligned}
\end{equation}
where $\oplus$ denotes the vector concatenation.
Similarly, embedding vectors of position $o\in [1,T]$ and response $r\in \{0,1\}$ are $\mathbf{e}_o=\textbf{Emb}^O_o\in\mathbb{R}^d$ and $\mathbf{e}_r = \textbf{Emb}^R_r\in\mathbb{R}^d$, respectively.

\subsection{Generator}\label{sec:gen}
The generator $\mathcal{G}([(q_t, \hat r_t)]_{i=t}^{T-1}, q_{T})$ predicts the student's response to question $q_T$ based on historical questions with its own predicted responses.
We utilize Gated Recurrent Unit (GRU) to represent students' learning states at step $T$
\begin{equation}
   \mathbf{h}_T = \text{GRU}_g(\mathbf{x}_T,\mathbf{h}_{T-1}),
\end{equation}
where $\mathbf{x}_T$ is the student interaction vector in step $T$ which would be introduce later. 
According to~\cite{EERNNA}, the student's current state is a weighted sum aggregation of all the historical states.
Hence, we assign different levels of importance to the historical states by applying an attention layer:
\begin{align}
     \mathbf{Q}=\mathbf{v}_{q_T}, \mathbf{K} &= \{\mathbf{v}_{q_1},\cdots, \mathbf{v}_{q_{T-1}}\}, \mathbf{V} = \{\mathbf{h}_1, \cdots, \mathbf{h}_{T-1}\},\nonumber\\
    \mathbf{a}_T &= \text{Attention}(\mathbf{Q}, \mathbf
    {K}, \mathbf{V}).
\end{align}

Next, we apply a prediction layer to calculate the probability of answering question $q_T$ correctly:
\begin{equation}
    \hat{r}_T = \text{Sigmoid}\left(\text{PredLayer}_g(\mathbf{v}_{q_T} \oplus\mathbf{h}_{T-1} \oplus \mathbf{a}_T)\right),
\end{equation}
where PredLayer is a two-layer fully connected network.

Finally, according to the predicted student response $\hat{r}_T$, we concatenate the question representation $\mathbf{v}_{q_T}$ and weighted historical state $\mathbf{a}_T$ to form the student interaction vector $\mathbf{x}_T$, which serves as the recurrent input for the next GRU process:
\begin{equation}
\begin{aligned}
    &\mathbf{x}_T = \left\{ \begin{array}{ll}
         \mathbf{v}_{q_T} \oplus \mathbf{a}_T \oplus \mathbf{0_q} \oplus \mathbf{0_a}, & \hat{r}\ge 0.5,\\
        \mathbf{0_q} \oplus \mathbf{0_a} \oplus \mathbf{v}_{q_T} \oplus \mathbf{a}_T, & \hat{r}<0.5,
    \end{array}\right. 
\end{aligned}
\end{equation}
where $\mathbf{0_q}$ and $\mathbf{0_a}$ are zero vectors with the same size as the vector $\mathbf{v}_{q_T}$ and $\mathbf{a}_T$, respectively. 

\subsection{Discriminator}
The discriminator $\mathcal{D}\left([(q_t, r_t)]_{t=1}^T\right)$ distinguishes whether the sequence is derived from real data (\ie, positive) or fake data (\ie, negative). 
The responses in the input sequence can be either ground-truth responses $r_i$ or generated ones $\hat{r}_i$. 
For simplicity, hereinafter, we use $r_i$ to denote the input responses.
We adopt the classical multi-head attention architecture to encode the sequence, which can be formulated as:
\begin{equation}
\begin{gathered}
    \text{MultiHead}(\mathbf{Q},\mathbf{K},\mathbf{V}) = \left(\mathbf{head}_1\oplus \dots\oplus \mathbf{head}_h\right) \mathbf{W}^o, \\
   \mathbf{head}_i = \text{Attention}\left(\mathbf{W}_i^q \mathbf{Q}, \mathbf{W}_i^k\mathbf{K}, \mathbf{W}_i^v \mathbf{V}\right), 
\end{gathered}
\end{equation}
where the number of heads $h$ is a hyperparameter, $\mathbf{W}_i^q, \mathbf{W}_i^k, \mathbf{W}_i^v$ and $\mathbf{W}^o$ represent the learnable parameters.

The discriminator takes the question sequence $[q_i]_{i=1}^T$, response sequence $[r_i]_{i=1}^T$, and positions $[o_i]_{i=1}^T$ as inputs. 
We define the question-position matrix $\mathbf{v}_q$ and response-position matrix $\mathbf{v}_r$ as:
\begin{equation}
    \mathbf{V}_q =\left(\begin{array}{c}
         \mathbf{e}_{q_1} \oplus \mathbf{e}_{o_1}  \\
          \mathbf{e}_{q_2} \oplus \mathbf{e}_{o_2}  \\
         \cdots \\
         \mathbf{e}_{q_T} \oplus \mathbf{e}_{o_T}
    \end{array}\right),\;
    \mathbf{V}_r = \left(\begin{array}{c}
         \mathbf{e}_{r_1} \oplus \mathbf{e}_{o_1}  \\
          \mathbf{e}_{r_2} \oplus \mathbf{e}_{o_2}  \\
         \cdots \\
         \mathbf{e}_{r_T} \oplus \mathbf{e}_{o_T}
    \end{array}\right).
\end{equation}
Then, we apply multi-head attention module to encode the sequence and finally output the estimated score of the sequence:
\begin{equation}
\begin{aligned}
    \mathbf{w}_q &= \text{MultiHead}_q(\mathbf{V}_q, \mathbf{V}_q, \mathbf{V}_q),\\
   \mathbf{w}_r &= \text{MultiHead}_r(\mathbf{V}_r, \mathbf{V}_r, \mathbf{V}_r),\\
    \textbf{D}_o &= \text{MultiHead}_d(\mathbf{w}_q, \mathbf{w}_r, \mathbf{w}_q\oplus \mathbf{w}_r),\\
    score &= \text{Sigmoid}(\text{PredLayer}_d(\textbf{D}_o)).
    \end{aligned}
\end{equation}
The structure of $\text{PredLayer}_d$ is identical to $\text{PredLayer}_g$.
The final output $score\in [0,1]$ denotes the probability that the input sequence is a positive sample.

\subsection{Adversarial Training}\label{sec:ad-train}
In AdvKT, the generator and the discriminator are alternately updated according to their respective loss functions until convergence. 
Meanwhile, data augmentation enriches training data to enhance the discriminator's robustness, enabling it to guide the generator in simulating realistic responses through adversarial training.

\subsubsection{Generator Loss}
We train the generator in a multi-step paradigm to minimize the gap between training and inference.
The generator has three training goals:  (1) aligning predictions with real-world data, (2) making the generated data indistinguishable to the discriminator, and (3) learning more robust and generalized question representations.

The binary cross-entropy (BCE) loss provides a supervised loss to align with the real-world data:
\begin{equation}
\label{BCELoss}
    \mathcal{L}_{\text{BCE}} = -\sum\nolimits_{t=1}^T\left(r_t \log(\hat r_t) + (1-r_t) \log(1-\hat r_t)\right).
\end{equation}
After generating the responses, the generator obtains a score from the discriminator, and can be adversarially updated by policy gradient~\cite{sutton2000policy}:
\begin{equation}
\label{RLLoss}
    \mathcal{L}_\text{Adv} = \frac{1}{T}\sum\nolimits_{t=1}^T |\hat r_t-0.5|\times R_{t} ,
\end{equation}
where $|\hat{r}_t-0.5|$ is the confidence, and $R_{t}$ represents the accumulative reward as follows:
\begin{equation}
\label{accu_reward}
    R_{t} = -\log\left(1- \mathcal{D}([(q_i,r_i)]_{i=1}^t\right) + \gamma R_{t+1},
\end{equation}
where $\gamma\in[0,1]$ is the decay factor.
Moreover, inspired by~\cite{9534358}, for more generalized question representations, we also introduce an additional autoregressive loss $\mathcal{L}_{AR}$ to predict the next question $q_t$ based on the student interaction vector $\mathbf{x}_T$:
\begin{equation}
\begin{aligned}
\label{LAR}
    \hat q_t &= \text{PredLayer}_{gq}(\mathbf{x}_T)\in\mathbb{R}^{|\mathcal{Q}|},\\
    \mathcal{L}_\text{G-AR} &= -\sum\nolimits_{t=1}^T\sum\nolimits_{i=1}^{|\mathcal{Q}|}\mathbb I(q_t=i)\log \hat q_t[i].
\end{aligned}
\end{equation}

Overall, the objective function of the generator is:
\begin{equation}
    \label{L-gen}
    \mathcal L_\text{G} =\mathcal L_\text{BCE}+ \lambda_1 * \mathcal L_\text{Adv} + \lambda_2 * \mathcal{L}_\text{G-AR},
    \end{equation}
where $\lambda_1$ and $\lambda_2$ are hyperparameters.

\subsubsection{Discriminator Loss}
Due to the data sparsity, we artificially enrich the positive and negative samples based on the student learning pattern to improve the robustness of the discriminator.

Positive samples comprise the original dataset $\mathcal R$ and augmented data $\mathcal T$.
The enrichment of positive samples includes four types: mask, crop, permute, and replace.
\begin{itemize}[leftmargin=10pt]
    \item \textbf{Mask}: Replace some questions in the origin sequences with a special token [MASK].
    \item \textbf{Crop}: Extract subsequences from the origin sequence.
    \item \textbf{Permute}: Randomly shuffle a subsequence.
    \item \textbf{Replace}: Calculate the difficulty of each question $q$ by $$\text{Difficulty}_q = \frac{\# \text{Correctly Answering } q}{\# \text{Answering } q},$$
    and randomly replace questions that the student answers correctly or incorrectly with easier or more difficult ones.
\end{itemize}
These four augmentation methods generally follow the student learning patterns, ensuring the rationality of the augmented data.

Negative samples include generative data $\mathcal E$ and label-reversed data $\mathcal{V}$. 
The generative data refer to the sequences with responses generated by the generator instead of the ground-truth ones. In addition to the vanilla sequences in the original dataset, we design a heuristic method to generate synthetic sequences to be further labeled by the generator.
Specifically, for each question pair $(q_A, q_B)$, we calculate the probability of question $q_B$ occurring after question $q_A$:
\begin{equation}
    p(q_A,q_B) = \frac{\# (q_A, q_B)}{\sum_{q\in \mathcal{Q}}\# (q_A, q)},\;\forall q_A, q_B\in\mathcal{Q}.
\end{equation}
We can sample the subsequent questions one by one based on the calculated probabilities above to generate more reasonable synthetic sequences. 
Furthermore, in addition to the generative data $\mathcal{E}$, we also selectively reverse the binary responses of the sequence in the original dataset $\mathcal{R}$ to create negative samples, denoted $\mathcal{V}$.

The training of the discriminator, based on the enriched positive and negative samples, aims to achieve two objectives: (1) to distinguish between positive and negative samples, and (2) to ensure the stability of the training process.
For sequence discrimination, the goal is to maximize the difference between the scores of each positive and negative sample. This is achieved by optimizing the following loss term:
\begin{equation}
    \mathcal L_\text{dist} = -\frac{1}{|\mathcal R \cup \mathcal T||\mathcal G \cup \mathcal{V}|} \sum_{i\in \mathcal R \cup \mathcal T} \sum_{j\in \mathcal G \cup \mathcal{V}}(score_i - score_j).
\end{equation}
Moreover, to ensure the training stability, we introduce the Gradient Penalty proposed in~\cite{10.5555/3295222.3295327} to restrict the gradient of the discriminator:
\begin{equation}
     GP = \alpha \times (\left\|\nabla_{D_o} \mathcal D(q,r)\right\|_2-1)^2,
\end{equation}
where $\alpha$ is a hyperparameter to balance the gradient penalty.
In conclusion, the objective of the discriminator is:
\begin{equation}
\label{L-disc}
    \mathcal L_\text{D} = -\mathcal L_\text{dist} +  GP.
\end{equation}

\subsubsection{Adversarial Learning}
The general training algorithm is illustrated in Appendix~\ref{alg:AdvKT}. We employ an alternating update strategy to iteratively update the discriminator and the generator. When updating the discriminator, we first generate four types of training data with rules, and compute the loss function in Eq.~(\ref{L-disc}), through which the augmented discriminator learns gain robust capabilities.
 During the training of the generator, it receives rewards from the discriminator, guiding it to simulate more realistic responses, while also being trained by the additional losses in Eq.~(\ref{L-gen}).

Note that the augmented data and label-reversed data are not directly used as the training corpus for the generator, as their sequence properties are highly similar to those in the training set. This similarity limits the effectiveness of GRU-based models. However, it could enhance the attention mechanism by introducing diverse local variations and encouraging more detailed learning of key sequence features, making it more suitable for improving the discriminator.

\section{Experiment}

\subsection{Experimental Setup}
\subsubsection{Datasets}
To evaluate the performance of our model, we conduct experiments on four real-world public datasets: ASSIST09, EdNet, Slepemapy and Junyi. Datasets in this paper are from sampling real students' learning logs of different subjects, indicating that the students exhibit different latent learning patterns.
The detailed statistics of the datasets are shown in Table~\ref{tab:dataset}.
The datasets possess varying levels of packing density and distribution, enabling us to test our method's robustness across different data characteristics.
We retain students in the dataset who had more than $10$ interactions with the platform and select the last $200$ records for each student.
In order to simulate the scenario with limited data, we randomly select $500$ or $6000$ students for training, while reserving $20\%$ of the original datasets for testing.

\begin{table}[tb]
    \centering
    \small
    \caption{Dataset Statistics}
    \label{tab:dataset}
    \renewcommand\arraystretch{0.8} 
    \setlength{\tabcolsep}{3mm} 
    \begin{tabular}{l r r r r}
        \toprule
        \textbf{Dataset} & \textbf{ASSIST09} & \textbf{EdNet} & \textbf{Slepemapy} & \textbf{Junyi} \\
        \midrule
        Subject & Math & English & Geography & Math \\
        \# Students & 500 & 500 & 6,000 & 6,000 \\
        \# Records & 41,741 & 34,262 & 553,797 & 532,139 \\
        \# Questions & 15,003 & 13,170 & 4,332 & 2,164 \\
        \# Concepts & 122 & 190 & 1,332 & 40 \\ \midrule
        Attempts per Q. & 2.78 & 2.60 & 127.8 & 245.90 \\
        Attempts per C. & 342.14 & 180.32 & 415.76 & 13,303.47 \\
        \bottomrule
    \end{tabular}
\end{table}

\subsubsection{Baselines and Evaluation Metrics}
To evaluate the effectiveness of our model, we compare AdvKT with 15 frequently used KT models.  Models trained based on BCE loss include DKT~\cite{DKT}, DFKT~\cite{DFKT}, SAINT~\cite{SAINT}, EERNNA~\cite{EERNNA}, AKT~\cite{AKT}, CKT~\cite{CKT}, SAKT~\cite{SAKT}, GKT~\cite{GKT},  DKVMN~\cite{DKVMN}, SKVMN~\cite{SKVMN}, LBKT~\cite{LBKT}, and simpleKT~\cite{simplekt}.
Models trained with other objectives include DHKT~\cite{DHKT}, IEKT~\cite{IEKT} and CL4KT~\cite{CL4KT}.

For a fair comparison, these baseline models are also trained under a multi-step setting, \ie, the input of them are question sequences and historically predicted responses.
We adopt Accuracy (ACC) and the Area Under Curve (AUC) as metrics.
A higher AUC or ACC indicates better performance of the KT task.

\subsubsection{Implementation Detail}
The maximum length of each student's learning sequence is $200$.
The dimension of the hidden state of GRU is set to $64$.
The question embedding, concept embedding, response embedding and position embedding all have a dimension of $64$.
We utilize a $4$-headed Transformer in the discriminator.
The value of $\gamma$ in Eq.~(\ref{RLLoss}) is chosen in \{$0.9$, $0.93$, $0.95$ , $0.98$\}.
In Eq.~(\ref{L-gen}), $\lambda_1$ is set to $1000$, and $\lambda_2$ is chosen from $\{0,1\}.$
The optimizer used is Adam~\cite{Adam}.
We update the discriminator every $2$ epochs.
The learning rate of the generator is $0.001$, while the learning rate of the discriminator is $0.005$.
The memory overhead incurred during training by AdvKT is comparable to that of recent knowledge tracing models, with AdvKT requiring approximately 15,000 MiB, while LBKT~\cite{LBKT} utilizes around 12,600 MiB.
During inference stage, AdvKT only requires the use of the generator,  thereby reducing GPU memory requirements by half.
Our model is implemented on PyTorch and is available on Github\footnote{Source code for AdvKT: \url{https://github.com/fulingyue/AdvKT}}.

\begin{table*}[t]
\centering
\caption{Overall performance of AdvKT on four public datasets. The best performing and second-best models are denoted in bold and underlined. $^\ast$ indicates p-value $< 0.05$ in the significance test.}
\label{tab:experiment}
\renewcommand\arraystretch{1.0}
\setlength{\tabcolsep}{1mm}
\small 
\resizebox{\textwidth}{!}{  
\begin{tabular}{c l c c c cc cc cc c}
   \toprule
\multicolumn{1}{c}{\multirow{2}{*}{{\textbf{Groups}}}} &\multicolumn{1}{c}{\multirow{2}{*}{{\textbf{Models}}}} &  \multicolumn{2}{c}{\textbf{ASSIST09}}  &  \multicolumn{2}{c}{\textbf{EdNet}} & \multicolumn{2}{c}{\textbf{Slepemapy}} & \multicolumn{2}{c}{\textbf{Junyi}} & \multicolumn{2}{c}{\textbf{Average Performance}}\\
\cmidrule(r){3-4} \cmidrule(lr){5-6} \cmidrule(lr){7-8} \cmidrule(lr){9-10}\cmidrule(r){11-12}
& & \textbf{ACC} & \textbf{AUC} &\textbf{ACC} & \textbf{AUC} & \textbf{ACC} & \textbf{AUC} & \textbf{ACC} & \textbf{AUC} & 
 \textbf{Avg. ACC} & \textbf{Avg. AUC}\\
\cmidrule{1-12}
\multirow{12}{*}{\textbf{BCE Loss}}& AKT & 0.6689 & 0.6178 & 0.6318& 0.6515 &  0.7611 & 0.6520 & \underline{0.7532} & 0.7547  & \underline{0.7037} &\underline{0.6690}\\
& SAINT & 0.6344 & 0.5272 &0.6166 &0.6384 &  0.7605 & \underline{0.6553} &0.7355 & \underline{0.7599} &0.6867 & 0.6452\\
& SAKT & 0.6127 & 0.5513 & 0.6134 & 0.6333& 0.7587 & 0.6393 &  0.7319 & 0.7525 & 0.6792 & 0.6441\\
& CKT &   0.6813 & 0.6342 & 0.5976 & 0.5959 &  0.6884 &0.5643 & 0.7082 & 0.6930 & 0.6689 & 0.6218\\
& DFKT & 0.6426& 0.6202  & 0.6331& 0.6589&0.6959& 0.5902 &0.7291&0.7394 & 0.6751 & 0.6522 \\
& DKT & 0.6411 & 0.6188 & 0.6297&0.6525 &0.7474 &0.6171 & 0.7289 & 0.7360 & 0.6868 & 0.6561\\
& EERNNA & \underline{0.6913} &  0.6182 & 0.6340&0.6341 & 0.7535 &0.6281  &0.7076 & 0.7161 & 0.6966 & 0.6491\\
& SKVMN & 0.6833 & 0.5515 & 0.6332 &0.6289 &0.7428& 0.6063 & 0.7126 & 0.7334 & 0.6930 & 0.6300\\
& DKVMN & 0.6823 & 0.5323 &0.6293 &0.6391&\underline{0.7609} &0.6049  & 0.7229 & 0.7265 & 0.6988 & 0.6257\\
& GKT &  0.6728&  0.5861 & 0.6001&0.5819 & 0.6994&0.6039 & 0.6786 & 0.6228 & 0.6627 & 0.5986\\
& LBKT & 0.6857&	0.6249	&0.5961&	\underline{0.6595}&	0.7603&  0.6021	&0.6807&0.7108 & 0.6807 &  0.6493 \\
& simpleKT &0.6741&	\underline{0.6402}&	0.5925	&0.6151&	0.6976 & 0.5810 	&0.7079& 0.7318&  0.6680 &0.6420 \\
\midrule
\multirow{3}{*}{\textbf{Other Objectives}}& DHKT & 0.6898 & 0.6224 &0.6310 & 0.6303& 0.7393& 0.6119  &  0.7167 & 0.7278 & 0.6942 & 0.6481 \\
& IEKT &0.6821 & 0.6260 & \underline{0.6376} & 0.6469 &  0.6908 &	0.5853 &  0.7188 & 0.7191  &  0.6823 & 0.6443 \\
& CL4KT & 0.6691 & 0.6061 & 0.6151 & 0.6268 & 0.7465& 0.6295 & 0.7303& 0.7578 & 0.6902 & 0.6550 \\
\midrule
\textbf{Adv. Learning}& AdvKT & \textbf{0.6993$^\ast$} & \textbf{0.6529$^\ast$} & \textbf{0.6464$^\ast$} &\textbf{0.6710$^\ast$}& \textbf{0.7642$^\ast$} &\textbf{0.6724$^\ast$} &\textbf{0.7552$^\ast$}  &\textbf{0.7863$^\ast$} & \textbf{0.7163$^\ast$} &\textbf{0.6956$^\ast$}\\
  \bottomrule
\end{tabular}
}
\end{table*}
\subsection{Overall Performance}
We compare AdvKT with all baselines under the multi-step training and inference setting. As shown in Table~\ref{tab:experiment}, we can obtain the following observations:
(1) Baseline models  have erratic performance across different datasets. 
    Due to the complexity of student response behaviors, existing models struggle to generalize well when confronted with datasets that emphasize different learning patterns.
    As a result, each model could only capture specific learning patterns and data characteristics, suffering from multi-step error accumulation and data sparsity. 
 (2) Baselines trained by BCE loss and other objectives show no significant differences compared to those trained with other objectives, because the effectiveness of alternative objectives depends on data distribution and can sometimes introduce noise.
(3) AdvKT generally achieves significant performance improvements over the baseline models in both ACC and AUC.
    On average, our model outperforms the best baseline (AKT) by $1.79\%$ on ACC and $3.98\%$ on AUC. The performance validates the effectiveness of our proposed AdvKT. 
    The adversarial learning paradigm not only help AdvKT acquire students' learning pattern under multi-step prediction scenarios, but also alleviate the data sparsity problem.

\begin{figure}[t]
    \centering
\subfigure{\includegraphics[width=0.35\linewidth]{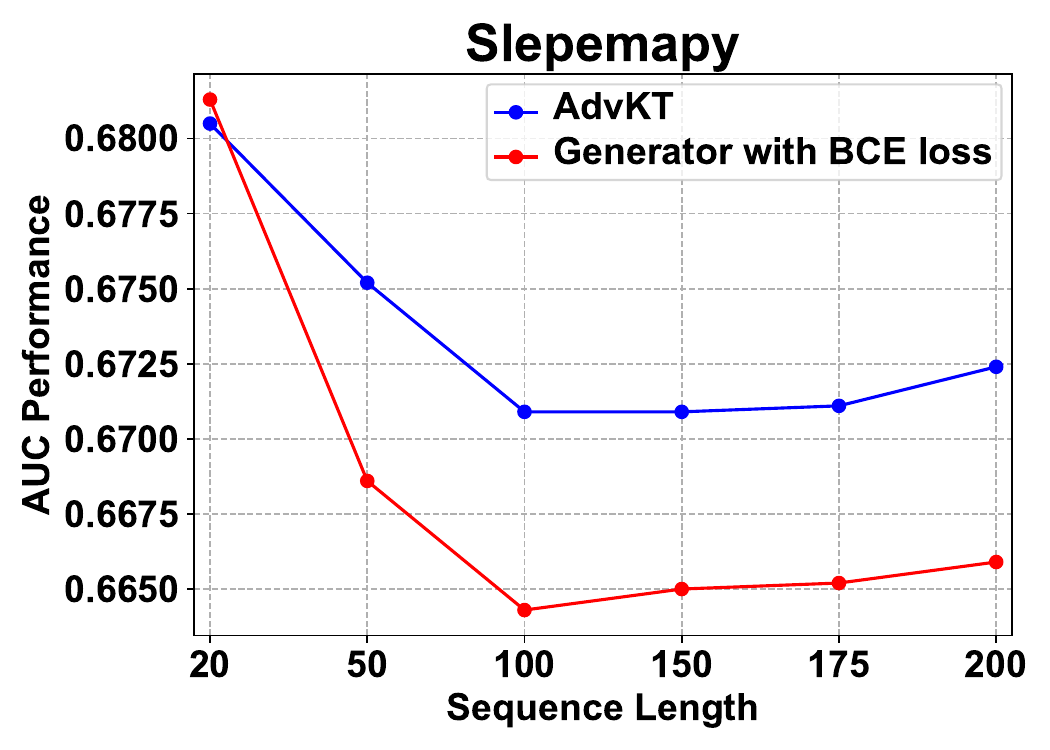}
    }
    \subfigure{
\includegraphics[width=0.35\linewidth]{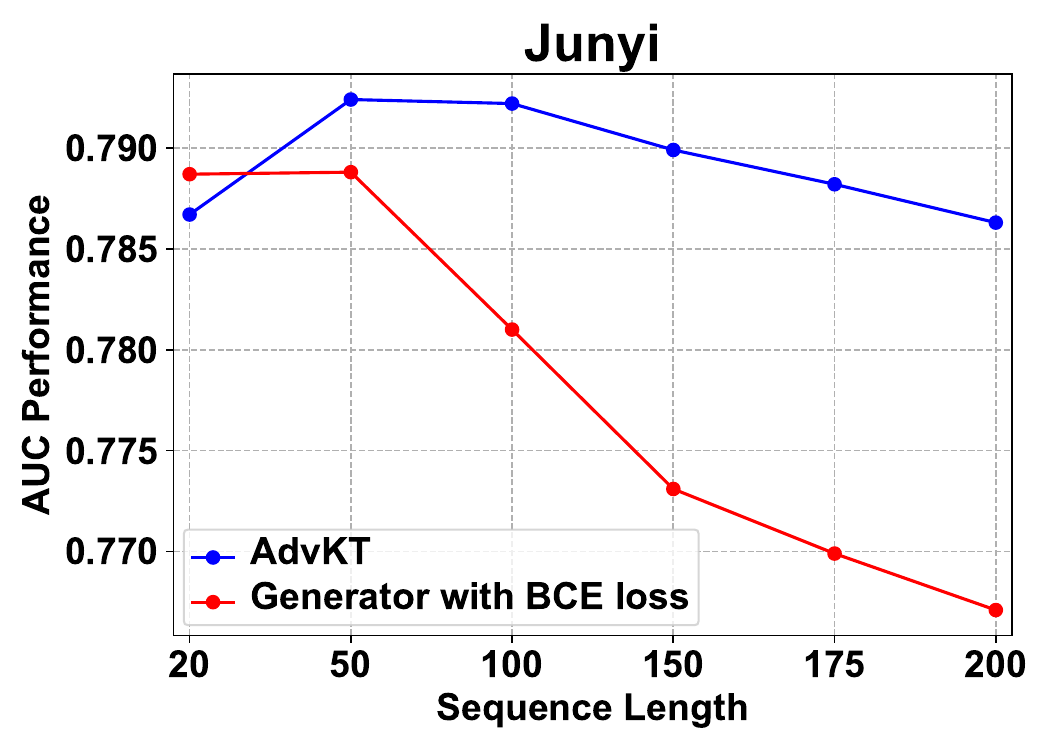}
    }
    \caption{Comparison results of the influence of learning sequence length of the generator with BCE loss and AdvKT.}
    \label{fig:multi-step}
\end{figure}
\begin{figure}[t]
\centering
\subfigure{\includegraphics[width=0.24\linewidth]{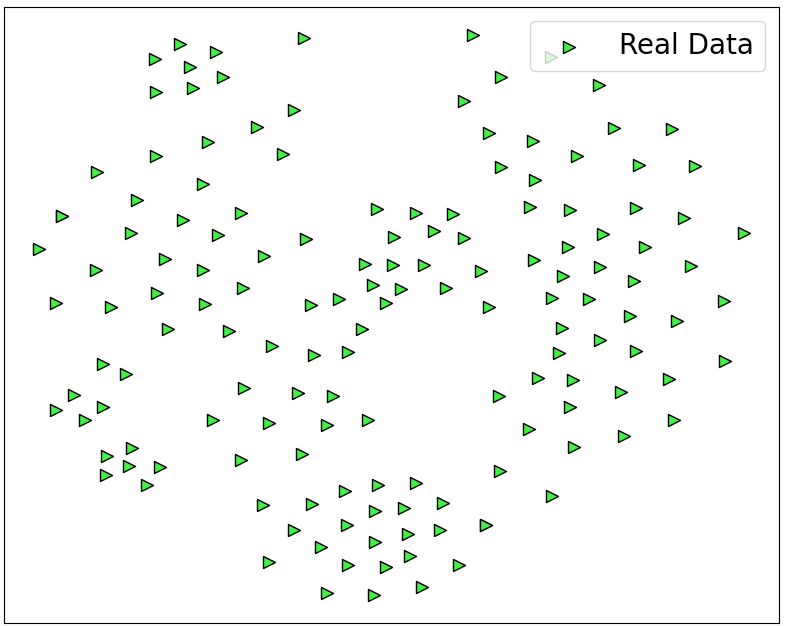}}
\subfigure{\includegraphics[width=0.24\linewidth]{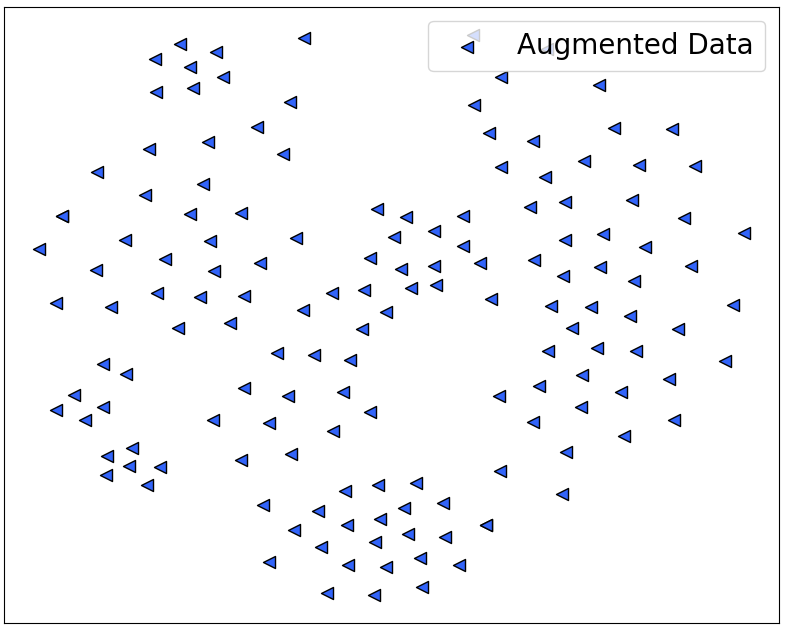}}
\subfigure{\includegraphics[width=0.24\linewidth]{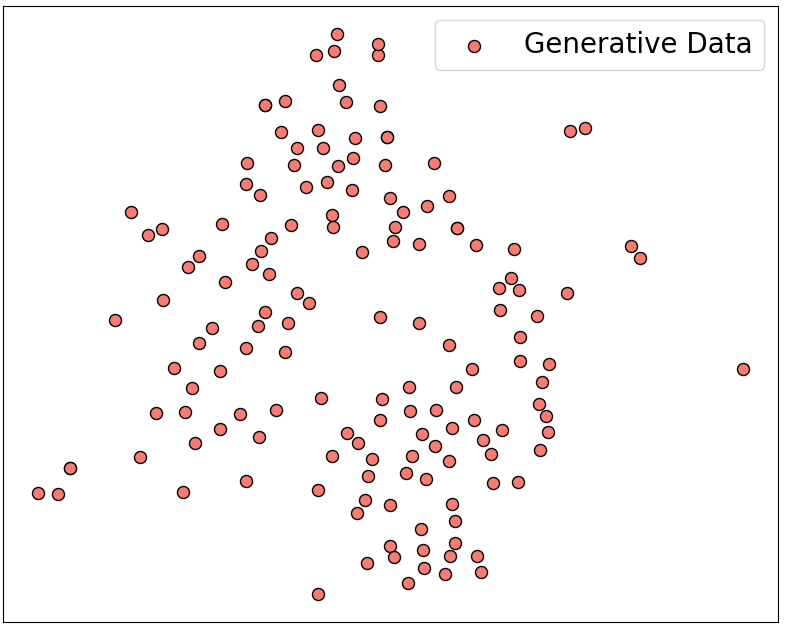}}
\subfigure{\includegraphics[width=0.24\linewidth]{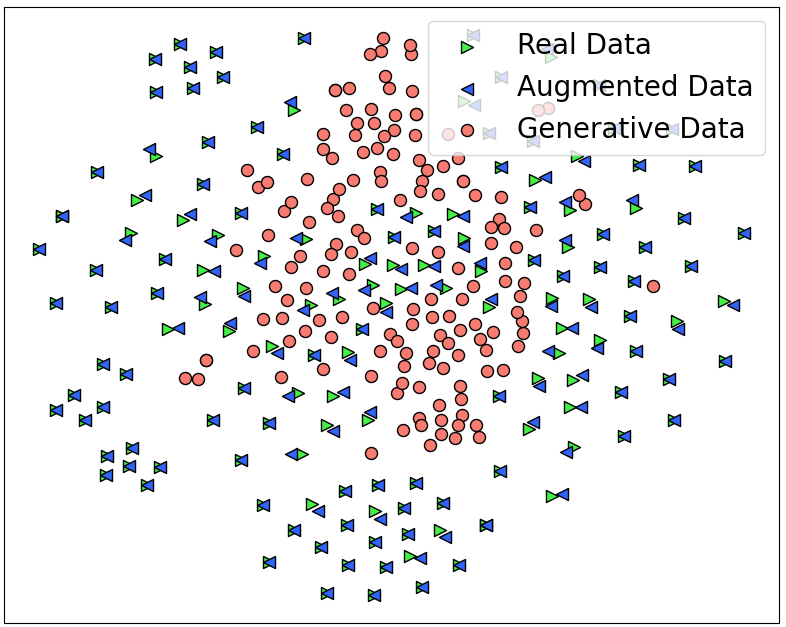}}

\caption{ Distribution of question-response pair visualized with two-dimensional t-SNE on ASSIST09. Green points are real data in dataset. Blue points correspond to data generated by positive data augmentation methods. Red points represent generative data.}
\label{fig:vis}
\end{figure}

\subsection{Mitigation of Error Accumulation}
In the context of multi-step prediction, error accumulation is a prevalent issue. 
Typically, in the prediction of long sequences, errors from previous predictions tend to have a more significant impact on the current step. 
We compare the performance of the generator trained by BCE loss only and our proposed AdvKT under various learning sequence lengths, with AUC as the metric.
As depicted in Figure~\ref{fig:multi-step}, our proposed AdvKT enjoys a relatively stable performance with different lengths of sequence, while the generator trained with BCE loss only suffers from severe error accumulation and meets a dramatic performance degeneration as the sequence length gradually increases. 
The comparison demonstrates the effectiveness of our proposed adversarial learning framework in AdvKT to tackle the error accumulation problem under real-world multi-step prediction scenarios.
The generator does not only fit the one-step data distribution via BCE loss, but also expand its horizon upon the entire sequence by receiving sequence-wise reward guidance from the discriminator.

\subsection{Mitigation of Data Sparsity}
In Figure~\ref{fig:vis}, we visualize the distribution of $(q,r)$ pairs in (\romannumeral1) original data $\mathcal R$ in the dataset, (\romannumeral2) augmented data $\mathcal{T}$ and (\romannumeral3) generative data $\mathcal{E}$ on ASSIST09. Visualization of distribution on other datasets are demonstrated in Appendix B.
 Our discriminator encodes the entire sequence and obtains hidden state representations. These representation vectors are reduced to two dimensions using t-SNE~\cite{van2008visualizingt_tsne}.

 The figure shows that the augmented data closely resembles the original data, validating the augmentation methods. However, there are noticeable gaps in both the original and augmented data, and the distribution boundaries are uneven. Although the augmented data increases the data size, it is insufficient to simply combine it with the original data for training a KT model. By generating new sequences, the generative data (red points) fills the gaps and boundary areas of the original data, ensuring a uniform distribution in the problem sequence subspace. Consequently, the discriminator can be trained more stably with the expanded dataset, providing better guidance for the generator to capture comprehensive student learning patterns. 

\subsection{Ablation Study}
\subsubsection{Ablation on Data Augmentation}
To investigate the contribution of data augmentation, we conduct ablation studies on four datasets.
We remove the following components to train the discriminator: (\romannumeral1) augmented data $\mathcal{T}$, (\romannumeral2) label-reversed data $\mathcal{V}$, and (\romannumeral3) both $\mathcal{T}$ and $\mathcal{V}$ simultaneously.
Note that the generative data $\mathcal{G}$ cannot be removed due to the requirement of adversarial learning.
The results, shown in Table~\ref{tab:abl-aug}, indicate that the performance of AdvKT decreases regardless of which type of data is removed. We attribute this to the fact that data augmentation provides more diverse data, enhancing the robustness of the discriminator and offering better reward guidance for the generator to capture student response patterns. Furthermore, augmented data contributes more significantly than label-reversed data, as it generates new sequences of questions.

   \begin{table}[tb]
    \centering
    \small
    \setlength{\tabcolsep}{3.5mm} 
    \renewcommand\arraystretch{0.9}
    \caption{Performance comparison between AdvKT and its variants: (\romannumeral1) train discriminator without augmented data $\mathcal{T}$; (\romannumeral2) train discriminator without reverse label data $\mathcal{V}$; (\romannumeral3) train discriminator without $\mathcal{T}$ and $\mathcal{V}$. Note that generative data cannot be removed, which is required by adversarial training.}
    \label{tab:abl-aug}
    \resizebox{\textwidth}{!}{  
    \begin{tabular}{l cccccccc}
        \toprule
        \multirow{2}{*}{\textbf{Models}} & \multicolumn{2}{c}{\textbf{ASSIST09}} & \multicolumn{2}{c}{\textbf{EdNet}} & \multicolumn{2}{c}{\textbf{Slepemapy}} & \multicolumn{2}{c}{\textbf{Junyi}} \\ 
        \cmidrule{2-9}
        & {ACC} & {AUC} & {ACC} & {AUC} & {ACC} & {AUC} & {ACC} & {AUC} \\ 
        \midrule
        AdvKT & \textbf{0.6993} & \textbf{0.6529} & \textbf{0.6464} & \textbf{0.6710} & \textbf{0.7642} & \textbf{0.6724} & \textbf{0.7552} & \textbf{0.7863} \\
        \midrule
        (\romannumeral1) w/o $\mathcal{T}$ & 0.6895 & 0.6486 & 0.6455 & 0.6693 & 0.7537 & 0.6664 & 0.7518 & 0.7836 \\
        (\romannumeral2) w/o $\mathcal{V}$ & 0.6916 & 0.6501 & 0.6462 & 0.6699 & 0.7564 & 0.6645 & 0.7521 & 0.7843 \\
        (\romannumeral3) w/o $\mathcal{V}$ \& $\mathcal{T}$ & 0.6887 & 0.6488 & 0.6460 & 0.6687 & 0.7557 & 0.6610 & 0.7544 & 0.7833 \\
        \bottomrule
    \end{tabular}
    }
\end{table}

  \begin{table}[t]
    \centering
    \caption{Ablation studies on different training loss: (\romannumeral 1) Remove $\mathcal{L}_\text{Adv}$ and train the generator without reward guidance from the discriminator, \ie, supervised learning; (\romannumeral 2) Remove $\mathcal{L}_\text{BCE}$ and train the generator only under the supervisions from the discriminator; (\romannumeral 3) Train the discriminator without gradient penalty; (\romannumeral 4) Replace $\mathcal{L}_\text{dist}$ for the discriminator with BCE loss.}
     \label{tab:abla_loss}
 \renewcommand\arraystretch{0.8}
\setlength{\tabcolsep}{3.5mm}
\small
\resizebox{\textwidth}{!}{  
    \begin{tabular}{lcccccccc}
    \toprule
       \multicolumn{1}{c}{\multirow{2}{*}{\textbf{Models}}} &\multicolumn{2}{c}{\textbf{ASSIST09}} & \multicolumn{2}{c}{\textbf{EdNet}} & \multicolumn{2}{c}{\textbf{Slepemapy}} & \multicolumn{2}{c}{\textbf{Junyi}}\\ \cmidrule{2-9}
         & {ACC}&{ AUC}& {ACC}&{ AUC} &{ACC}&{ AUC} & {ACC}&{ AUC}\\ \midrule
        AdvKT &\textbf{0.6993}& \textbf{0.6529} & \textbf{0.6464}& \textbf{0.6710} &\textbf{0.7642} &\textbf{0.6724} & \textbf{0.7552} & \textbf{0.7863}\\ \midrule
        (\romannumeral1)  w/o $\mathcal{L}_\text{Adv}$ & 0.6859 &  0.6451&  0.6462 & 0.6672 & 0.7479 & 0.6659 &  0.7532& 0.7671\\
       (\romannumeral 2) w/o $\mathcal{L}_\text{BCE}$ & 0.5766 &  0.5299 & 0.5931 & 0.5575 & 0.7083 & 0.5661 & 0.7125 & 0.6588\\ \midrule
        (\romannumeral 3) w/o GP & 0.6735 & 0.6342 & 0.6511 & 0.6683 & 0.7600 & 0.6284 & 0.7541 & 0.7842\\
       (\romannumeral 4) $\mathcal{L}_\text{dist}\rightarrow \mathcal{L}_\text{BCE}^\text{dist}$ & 0.6908 & 0.6497 & 0.6434 & 0.6698 & 0.7545 & 0.6685 & 0.7463 & 0.7818\\
         \bottomrule
\end{tabular}
    }
\end{table}
\begin{figure}[t]
\centering
\subfigure{\label{fig:ass09-realvec}
\includegraphics[width=0.23\linewidth]{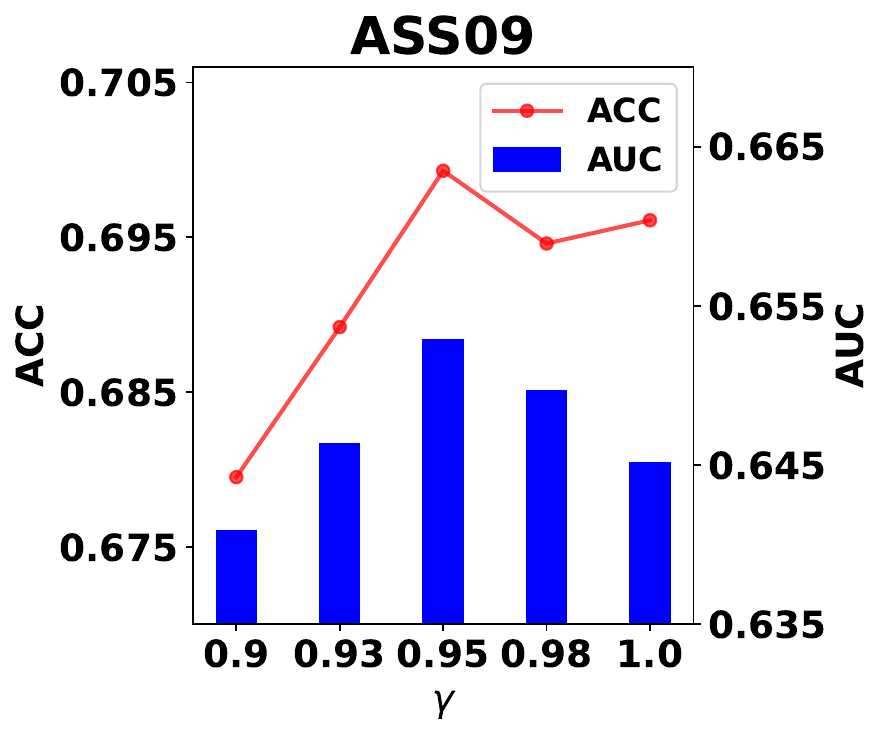}}
\subfigure{\label{fig:ass09-realvec}
\includegraphics[width=0.23\linewidth]{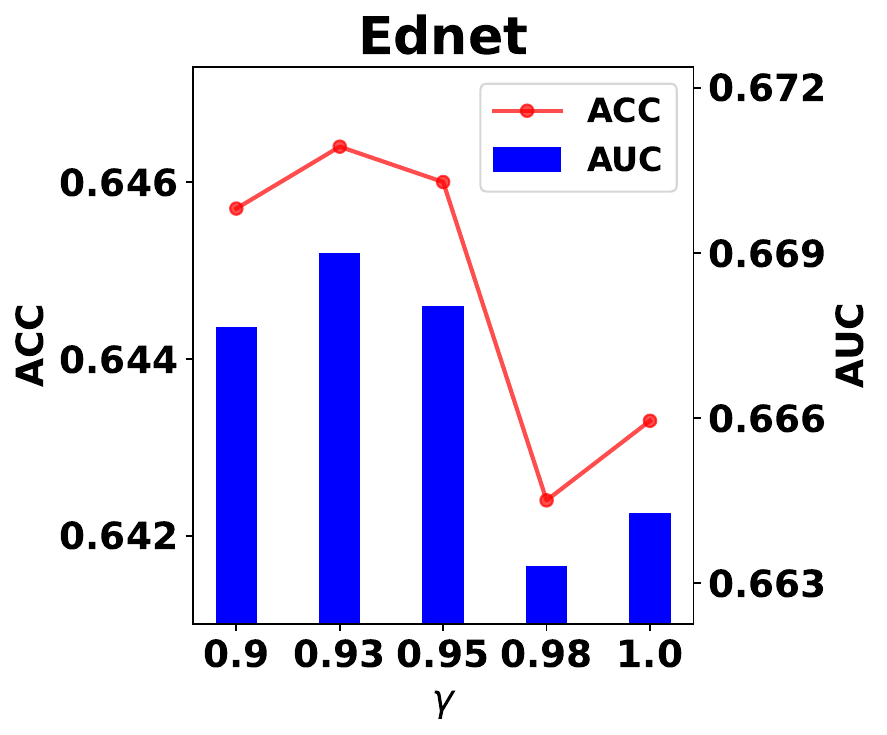}}
\subfigure{\label{fig:ass09-realvec}
\includegraphics[width=0.23\linewidth]{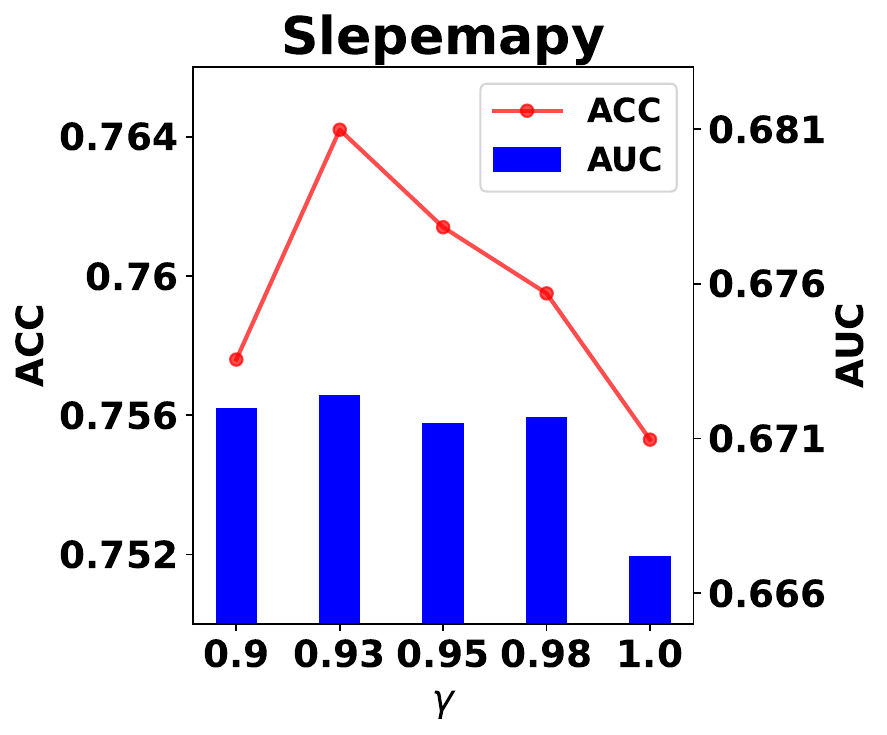}}
\subfigure{\label{fig:ass09-realvec}
\includegraphics[width=0.23\linewidth]{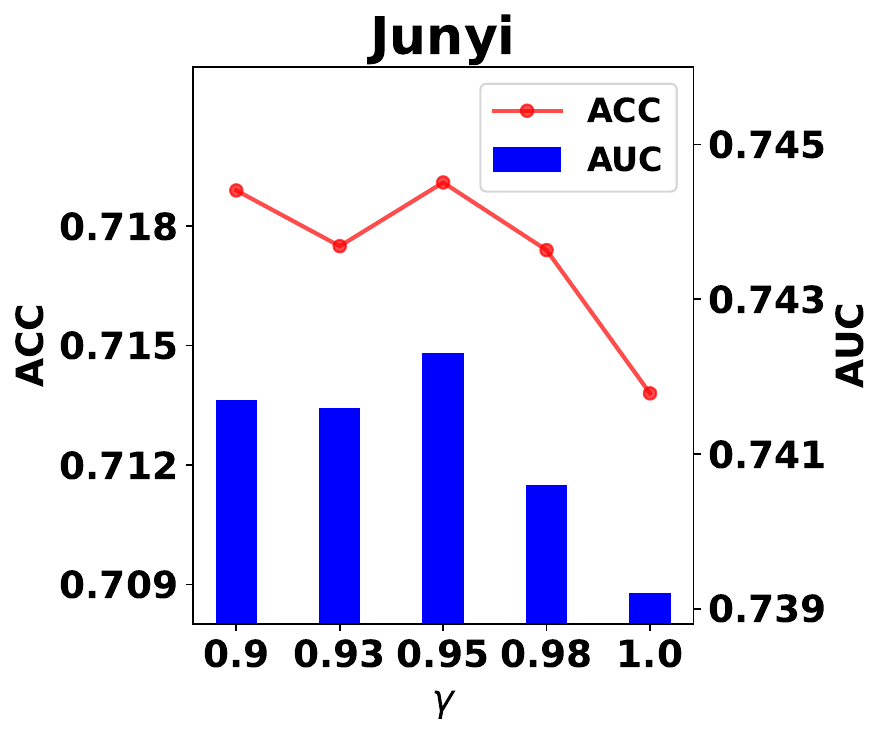}}
\caption{Ablation study on the hyper-parameter $\gamma$ on four datasets.}
\label{fig:abl-gamma}
\end{figure}

\subsubsection{Ablation on Loss Functions}
We investigate the impact of different loss functions designed for the generator $\mathcal{G}$ and the discriminator $\mathcal{D}$, with results presented in Table~\ref{tab:abla_loss}.

Firstly, we assess the contribution of loss functions for the generator. Removing either the adversarial loss $\mathcal{L}_\text{Adv}$ or the binary cross-entropy loss $\mathcal{L}_\text{BCE}$ significantly degrades AdvKT's performance. 
This suggests the necessity of balancing adversarial learning ($\mathcal{L}_\text{Adv}$) with supervised learning ($\mathcal{L}_\text{BCE}$) to effectively capture multi-step student response patterns.

Next, we perform an ablation study on the loss functions of the discriminator. 
We can observe that GP is crucial for maintaining the training stability of the discriminator, ensuring that the generator receives consistent and meaningful reward guidance for enhanced predictive performance. 
Moreover, replacing our custom-designed distance loss $\mathcal{L}_\text{dist}$ with a simple BCE loss $\mathcal{L}_\text{BCE}^\text{dist}$results in a noticeable performance decline for AdvKT.
The distance loss $\mathcal{L}_\text{dist}$ focuses more on the internal ranking of positive and negative samples, rather than merely achieving pointwise scores close to 0 or 1 as indicated by the BCE loss. Consequently, our distance loss assigns higher scores to positive samples than negative ones, ensuring more accurate reward guidance for the generator.

\subsubsection{Ablation on Discount Factor $\gamma$}
To assess the sensitivity of our model, we evaluate the impact of discount factor $\gamma$.
We test our method with $\gamma=\{0.9, 0.93, 0.95, 0.98, 1\}$ across four datasets, keeping other hyperparameters constant. 
As illustrated in Figure~\ref{fig:abl-gamma}, the performance fluctuates when tuning the discount factor.
The discount factor $\gamma$ determines the importance the generator places on future predictive performance and should be adjusted based on the specific dataset requirements.

\subsection{Insight of Adversarial Learning}
\begin{figure}[t]
    \centering
    \includegraphics[width=0.65\linewidth]{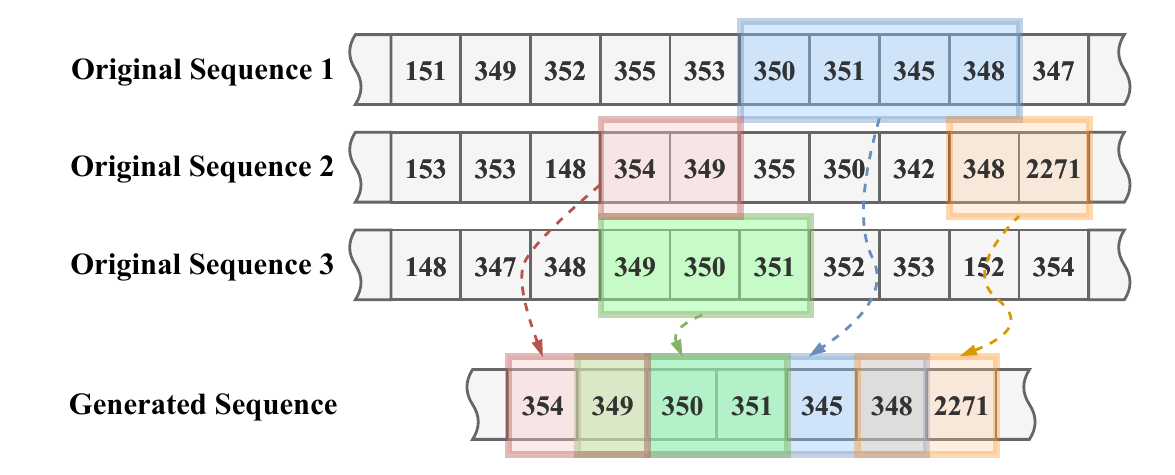}
    \caption{Demonstration of generated question sequences and original question sequences.}
    \label{fig:casestudy}
    
\end{figure}

Generative data $\mathcal{G}$ plays a crucial role in training the generator alongside the real dataset. 
The quality and diversity of these synthetic sequences directly influence the supervision signal  ($\mathcal{L}_\text{adv}$) obtained by the generator. In Figure~\ref{fig:casestudy}, we present a generated synthetic sequence along with its most similar learning sequences from the dataset.  Four subsequences are found within the original sequences, highlighting a certain similarity between the generated and original sequences. However, each generated sequence remains distinct from any single original sequence.

This similarity ensures that the generated sequences maintain the logical consistency of the original data, such as presenting simpler questions before more challenging ones. Meanwhile, the diversity in the generated sequences helps AdvKT tackle data sparsity by enriching the dataset with varied examples. This diversity also enables the model to learn from accumulated errors across different sequences, further reducing the impact of error accumulation.

\section{Related Work}
\subsection{Knowledge Tracing}
To investigate students' learning patterns, numerous knowledge tracing models have been proposed under a single-step training paradigm.
Deep Knowledge Tracing (DKT)~\cite{DKT} pioneers using the deep learning-based knowledge tracing methods.
It uses the hidden states of long short-term memory (LSTM)~\cite{LSTM} networks to describe the students' knowledge states.
After DKT, DHKT~\cite{DHKT} and other variants of DKT introduce extra information like question-skill relations and additional regularization terms to improve the performance. 
Inspired by Key-Value Memory Network (KVMN)~\cite{KVMN}, several works~\cite{DKVMN,SKVMN} use key-matrix and value-matrix to represent concepts and student's mastery level.
GKT randomly builds a similarity graph of skills, GIKT~\cite{GIKT} applies the graph of skills and questions to obtain better representations of questions.
After the transformer architecture was proposed, some works~\cite{SAINT,EERNNA} tried to utilize the attention mechanism in KT tasks. 

Recently, StableKT~\cite{stableKT} highlighted the presence of error accumulation in single-step KT scenarios. All the aforementioned methods experience more severe error accumulation under multi-step inference. At the same time, they suffer significant performance degradation when training data is sparse. Although approaches~\cite{guo2024mitigating,sinkt,MAMbakt} enhance learning from question text using large language models, educational datasets still face challenges at the ID level, with limited and uneven interaction data. AdvKT utilizes an adversarial learning paradigm, thereby improving both training paradigms and data utilization.

\subsection{Adversarial Learning}
In recent years, GAN~\cite{GAN} has been widely used for data augmentation in computer vision and Natural Language Processing.
Combining the idea of GAN with Inverse Reinforcement Learning (IRL), researchers propose GAIL~\cite{GAIL} framework to learn policy by imitating expert trajectories.
Many previous work~\cite{1911.03845,1805.10000,AICM} adopt GAN and GAIL frameworks for user modeling.
These work models and simulates users' behavior (clicking, buying) on the web search page or shopping websites.

Unlike user interest modeling, the KT task includes the change of users' state, \ie, knowledge state transition.
In AdvKT, we consider the students' learning patterns and generate logical question sequences to estimate the knowledge state of students. 

\section{Conclusion}
In this work, we propose a novel adversarial multi-step training framework for knowledge tracing (AdvKT), which, for the first time, explicitly models real-world multi-step prediction by introducing the adversarial learning framework.
AdvKT adversarially trains a generator and a discriminator, aiming at alleviating error accumulation and data sparsity problem.
The generator is designed to simulate students' learning process under the multi-step setting, \ie, generating students' responses based on the question sequence and its previous predicted responses instead of the ground-truth ones.
The discriminator, whose training data is augmented by well-designed rules, distinguishes whether the learning sequence is derived from real data or fake data, and therefore provides sequence-wise reward guidance for the generator to capture student response patterns.
Extensive experiments on four real-world datasets demonstrate the superiority of AdvKT compared with existing baseline models, as well as its capabilities for mitigating the error accumulation and data sparsity problem.

\begin{credits}
\subsubsection{\ackname} The Shanghai Jiao Tong University team is partially supported by National Key R\&D Program of China (2022ZD0114804), Shanghai Municipal Science and Technology Major Project (2021SHZDZX0102) and National Natural Science Foundation of China (62322603, 62177033).

\subsubsection{\discintname}
The authors have no competing interests to declare that are
relevant to the content of this article.
\end{credits}
\appendix
\input{appendix}
\end{document}

%% file: appendix.tex
\clearpage
\appendix
\section{Adversarial Training Algorithm}\label{alg:AdvKT}
We present an adversarial training paradigm to train the generator and discriminator together.
The overall training algorithm is illustrated in Algorithm~\ref{alg:advkt-training}.
We alternatingly update the discriminator and the generator until convergence.

\begin{algorithm}[h]
    \setstretch{1}
	\caption{Adversarial Training of AdvKT}
 \label{alg:advkt-training}
    \renewcommand\arraystretch{1.0}
    
	Initialize the Generator $\mathcal{G}$, Discriminator $\mathcal D$ randomly\;
 \While{\textnormal{not converged}}{
		
		\ForEach{batch data $\mathcal{R}$}{
            \uIf{Update Discriminator}{
            /* Update Discriminator */\\
                Generate augmented data $\mathcal{T}$;\\
                Generate generative data $\mathcal{E}$ with generator $\mathcal{G}$;\\
                Generate reversed label data $V$;\\
                Compute the loss function Eq.~(\ref{L-disc}) and update $\mathcal{D}$;
            }
            
            /* Update Generator */\\
            Compute the supervise learning loss Eqs.~(\ref{BCELoss}) and (\ref{LAR});\\
            Generate generative data $\mathcal{E}$;\\
			Compute the policy gradient loss  Eq.~(\ref{RLLoss}) using $\mathcal{E}$;\\
            Update $\mathcal{G}$ with Eq.~(\ref{L-gen});
		}
	
	}
\end{algorithm}

\begin{figure*}[t]
\centering
\subfigure{\label{fig:realvec}
\includegraphics[width=0.23\linewidth]{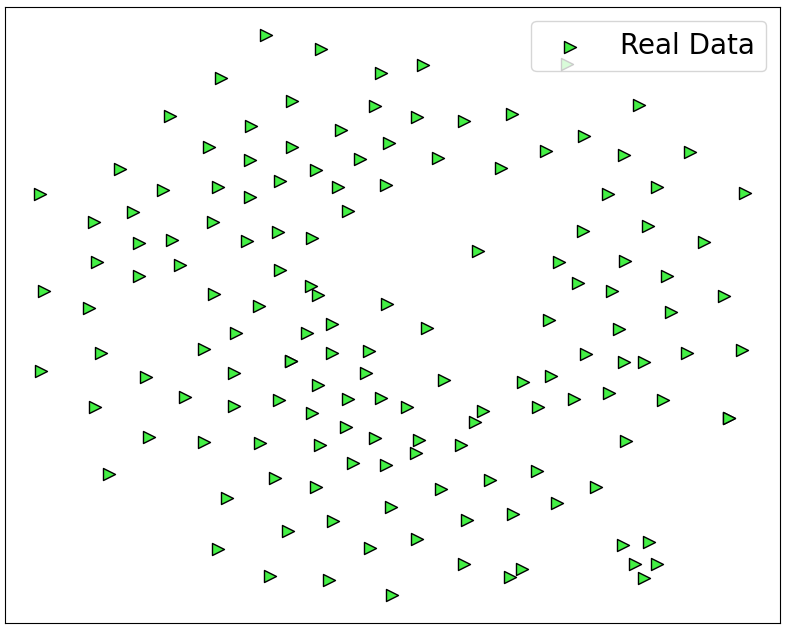}}
\subfigure{\label{fig:augvec}
\includegraphics[width=0.23\linewidth]{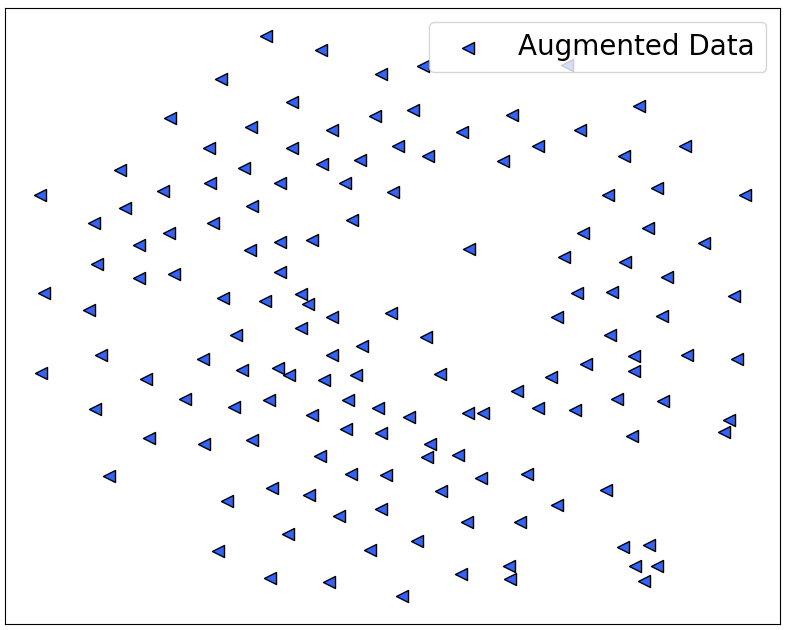}}
\subfigure{\label{fig:subfig:a}
\includegraphics[width=0.23\linewidth]{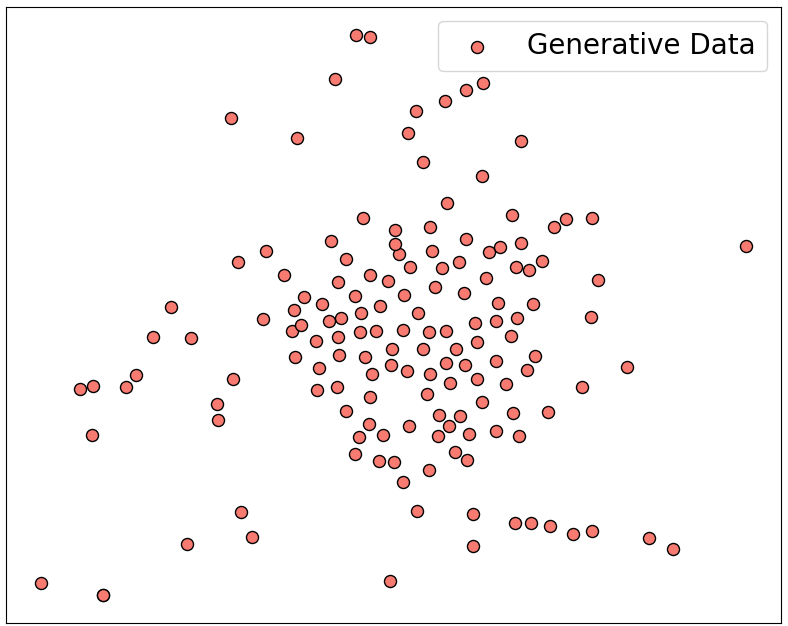}}
\subfigure{\label{fig:allvec}
\includegraphics[width=0.23\linewidth]{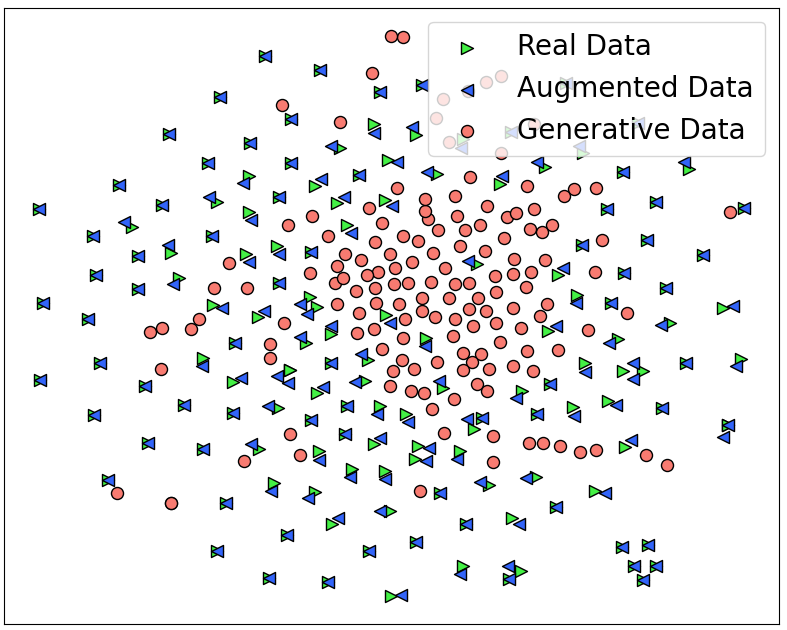}}
\\ \small EdNet

\subfigure{\label{fig:realvec}
\includegraphics[width=0.23\linewidth]{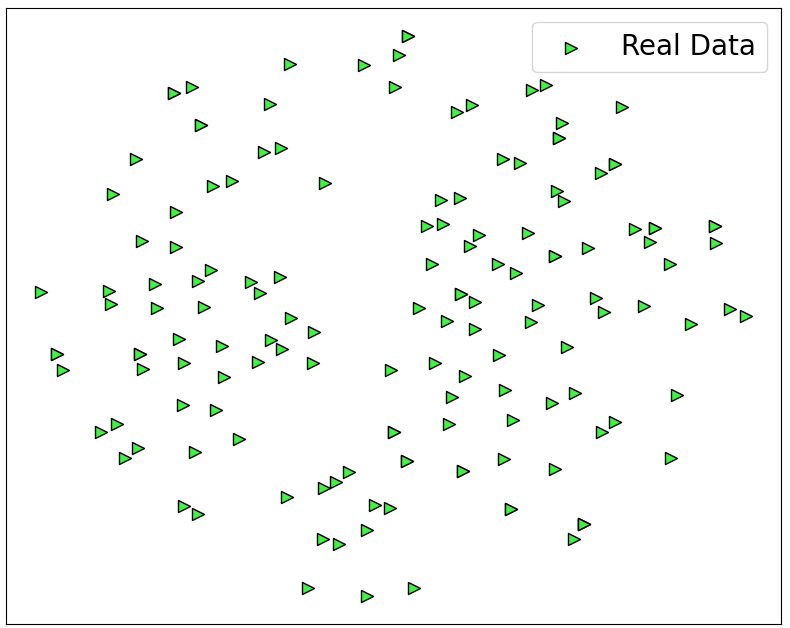}}
\subfigure{\label{fig:augvec}
\includegraphics[width=0.23\linewidth]{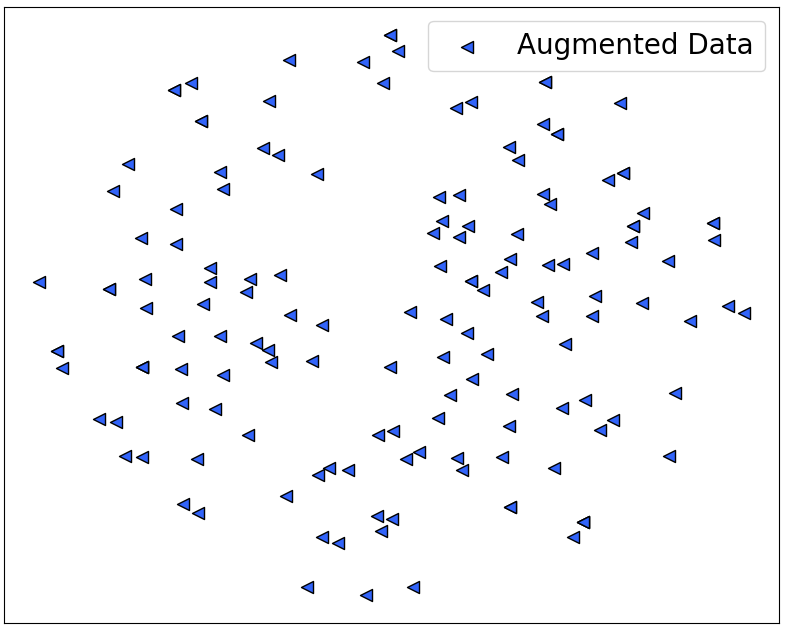}}
\subfigure{\label{fig:subfig:a}
\includegraphics[width=0.23\linewidth]{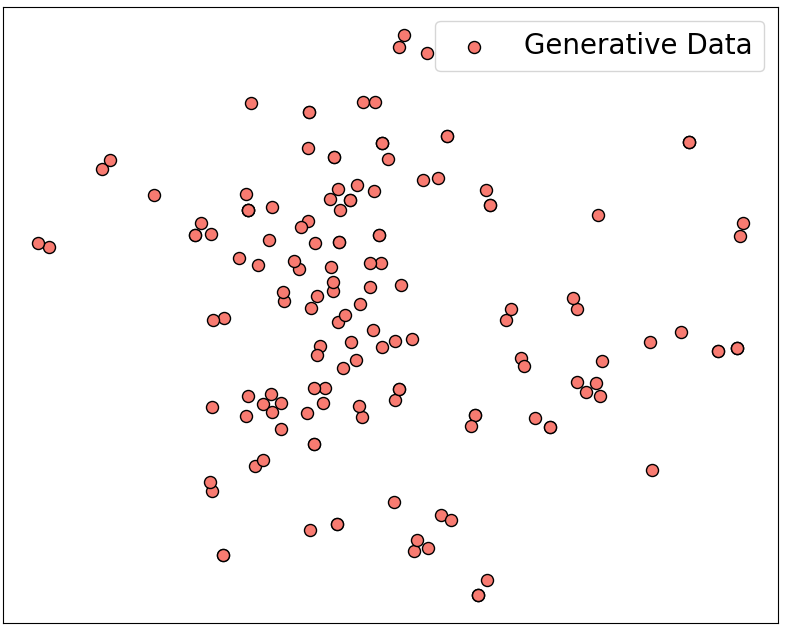}}
\subfigure{\label{fig:allvec}
\includegraphics[width=0.23\linewidth]{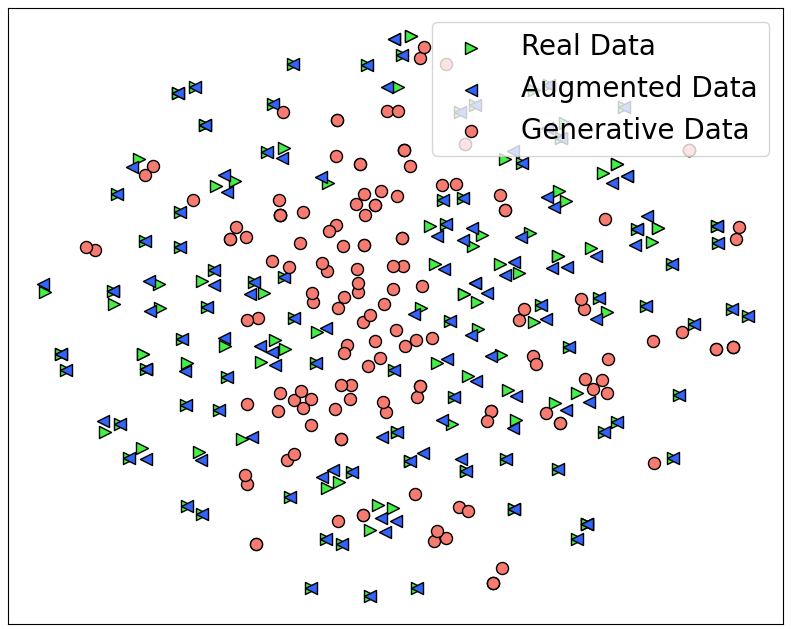}}
\\ \small  Slepemapy

\subfigure{\label{fig:realvec}
\includegraphics[width=0.23\linewidth]{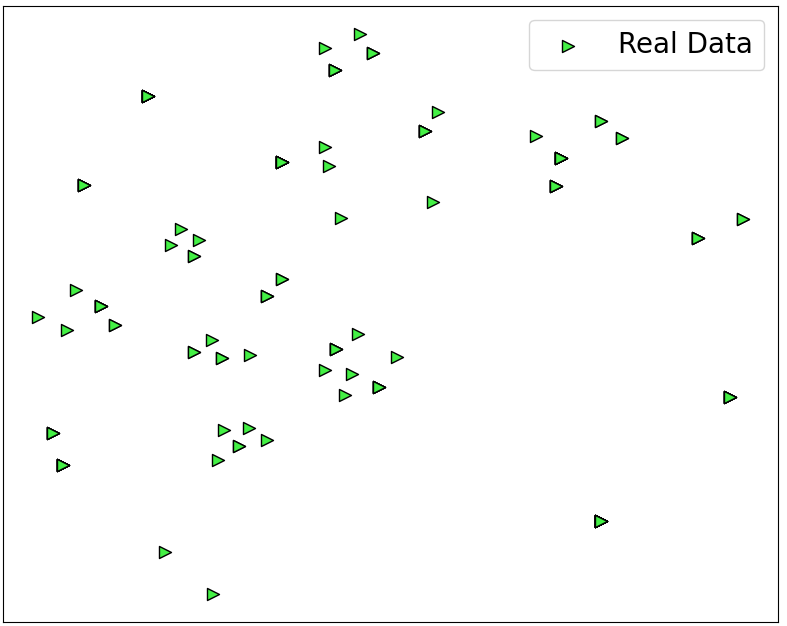}}
\subfigure{\label{fig:augvec}
\includegraphics[width=0.23\linewidth]{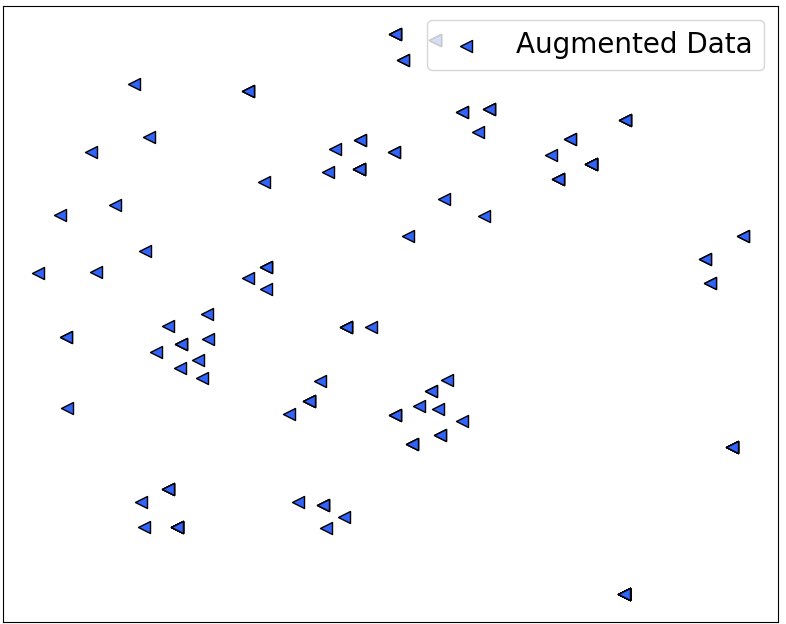}}
\subfigure{\label{fig:subfig:a}
\includegraphics[width=0.23\linewidth]{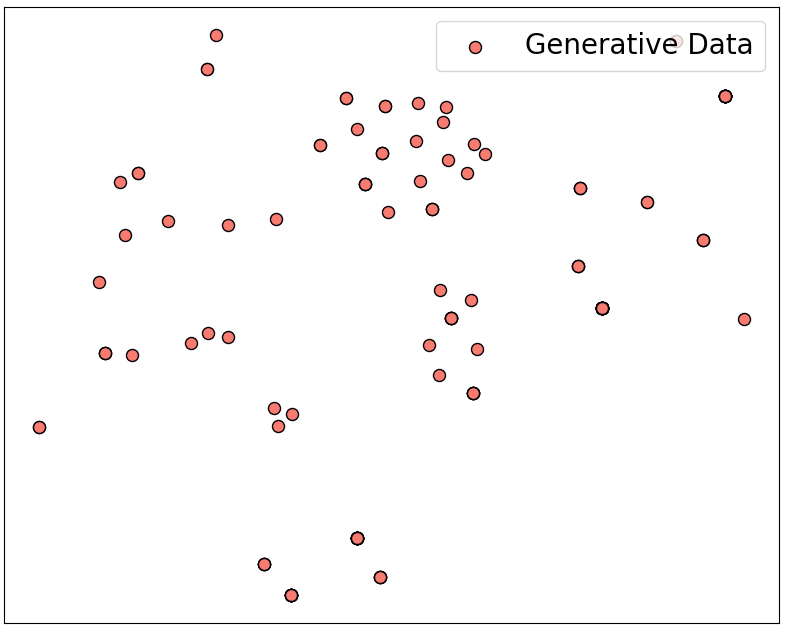}}
\subfigure{\label{fig:allvec}
\includegraphics[width=0.23\linewidth]{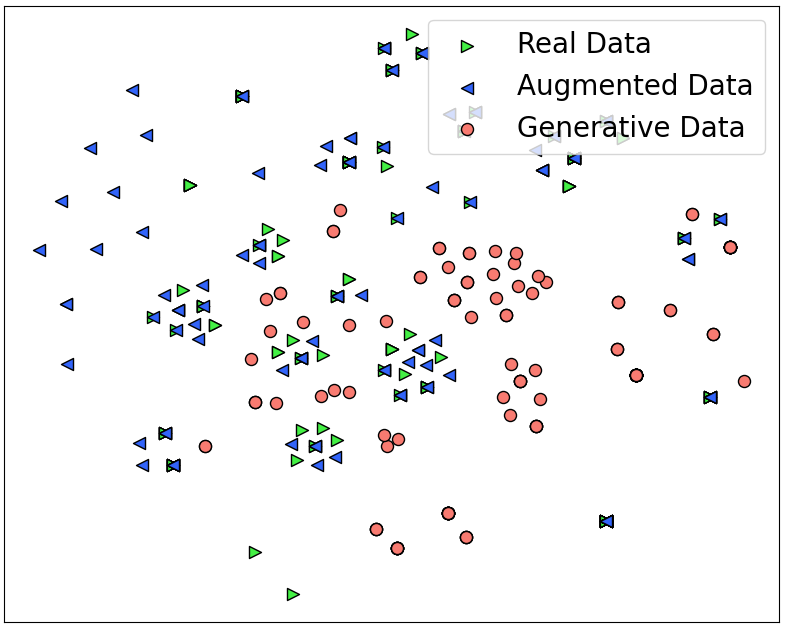}}
\\ \small Junyi
\caption{Distribution of question-response pair visualized with two-dimensional t-SNE. Green points are real data in dataset. Blue points correspond to data generated by popular data augmentation methods (mask, crop, permute, replace). Red points represent generative data. The three lines of images are visualizations of EdNet, Slepemapy and Junyi. }
\label{fig:app_vis}
\end{figure*}

\section{Data Distribution Visualization}\label{app:vis}
In Figure~\ref{fig:app_vis}, we visualize the distribution of $(q,r)$ pairs in (\romannumeral1) original data $\mathcal R$ in the dataset, (\romannumeral2) augmented data $\mathcal{T}$ and (\romannumeral3) generative data $\mathcal{E}$ on EdNet, Slepemapy and Junyi.
The figure illustrates that even though different datasets exhibit significant variations in data distribution, real data consistently suffer from the issues of data sparsity and non-uniformity. 
In EdNet's real data, there are three noticeable gaps, and the points along the data boundaries are also unevenly distributed. 
As for Slepemapy's real data, while the boundaries appear relatively uniform, there exists a distinct hole within.
Junyi's data, on the other hand, skews towards the  top left corner, with sparse points in the bottom right corner. 
Our four positive sample augmentation methods first expand the dataset with real data, ensuring that the augmented data still aligns with the real data. 
Additionally, generative data can fill in the gaps in real data, resulting in a more evenly distributed training dataset across the entire space.
Therefore, our data augmentation on the discriminator's training data could well solve the data unevenness and sparsity.